%% file: main.tex
\documentclass{article}

\usepackage[]{spconf}

\usepackage{amsmath,amssymb,amsfonts,amsthm,optidef}
\usepackage{algorithm,algpseudocode}
\usepackage{graphicx}

\usepackage{cite}
\usepackage{hyperref}
\usepackage[capitalise]{cleveref}
\newcommand{\letterref}[2]{\namecref{#1}~\hyperref[#1]{\ref*{#1}#2}}

\usepackage{color}

\usepackage{pgfplots}
\usepgfplotslibrary{groupplots}
\usepgfplotslibrary{colormaps}
\pgfplotsset{compat=1.16}
\pgfplotscreateplotcyclelist{mycyclelist}{%
  {red, mark=o, mark options={solid}, solid, thick},
  {blue, mark=x, mark options={solid}, dashed, thick},
  {orange, mark=*, mark options={solid}, dotted, thick},
  {violet, mark=o, mark options={solid}, dashed, thick},
  {red, mark=*, mark options={solid}, dashed, thick},
  {black, mark=x, mark options={solid}, solid, thick},
}
\pgfplotsset{cycle list name=mycyclelist}

\DeclareMathOperator{\diag}{diag}

\DeclareMathOperator*{\argmax}{arg\,max}
\DeclareMathOperator*{\argmin}{arg\,min}

\newcommand{\R}{\ensuremath{{\mathbb R}}}
\newcommand{\Rn}{\ensuremath{{\mathbb R}^n}}

\newcommand{\E}{\ensuremath{{\mathbb E}}}

\newcommand{\Nodes}{\ensuremath{{\mathcal V}}}
\newcommand{\Edges}{\ensuremath{{\mathcal E}}}
\newcommand{\Graph}{\ensuremath{{\mathcal G}}}

\newcommand{\Filt}{\ensuremath{{\mathcal H}}}
\newcommand{\Cy}{\ensuremath{{\mathbf C}_{\mathbf y}}}

\newcommand{\A}{\ensuremath{{\mathbf A}}}
\newcommand{\Shift}{\ensuremath{{\mathbf S}}}
\newcommand{\V}{\ensuremath{{\mathbf V}}}
\newcommand{\Vhat}{\ensuremath{\widehat{\mathbf V}}}
\newcommand{\U}{\ensuremath{{\mathbf U}}}
\renewcommand{\L}{\ensuremath{{\mathbf L}}}
\newcommand{\D}{\ensuremath{{\mathbf D}}}

\newcommand{\x}{\ensuremath{{\mathbf x}}}
\renewcommand{\v}{\ensuremath{{\mathbf v}}}
\newcommand{\vhat}{\ensuremath{\widehat{\mathbf v}}}
\renewcommand{\u}{\ensuremath{{\mathbf u}}}

\newcommand{\y}{\ensuremath{{\mathbf y}}}
\newcommand{\w}{\ensuremath{{\mathbf w}}}

\newcommand{\eg}{\textit{e.g.}}
\newcommand{\ie}{\textit{i.e.}}

\newcommand{\ER}{{Erd\H{o}s-R\'enyi}}
\newcommand{\BA}{{Barab\'asi-Albert}}

\newtheorem{problem}{Problem}

\newtheoremstyle{myremark}
{\topsep} 
{\topsep} 
{\normalfont} 
{} 
{\bfseries} 
{.} 
{5pt plus 1pt minus 1pt} 
{\thmname{#1}\thmnumber{ #2}\thmnote{ (#3)}} 

\theoremstyle{myremark}
\newtheorem{remark}{Remark}

\newcommand{\parheading}[1]{\vspace{1mm}\noindent\textbf{#1}}

\title{Network topology change-point detection from graph signals with prior spectral signatures}
\name{Chiraag Kaushik, T. Mitchell Roddenberry, and Santiago Segarra
  \thanks{This work was partially supported by NSF under award CCF-2008555.
    Emails: \href{mailto:cvk4@rice.edu}{cvk4@rice.edu}, \href{mailto:mitch@rice.edu}{mitch@rice.edu}, \href{mailto:segarra@rice.edu}{segarra.rice.edu}}}
\address{Rice University, Electrical and Computer Engineering, Houston, TX}

\begin{document}
\ninept
%
\maketitle
\begin{abstract}
  We consider the problem of sequential graph topology change-point detection from graph signals.
  We assume that signals on the nodes of the graph are regularized by the underlying graph structure via a graph filtering model, which we then leverage to distill the graph topology change-point detection problem to a subspace detection problem.
  We demonstrate how prior information on the spectral signature of the post-change graph can be incorporated to implicitly denoise the observed sequential data, thus leading to a natural CUSUM-based algorithm for change-point detection.
  Numerical experiments illustrate the performance of our proposed approach, particularly underscoring the benefits of (potentially noisy) prior information.
\end{abstract}
\begin{keywords}
Graph signal processing, change-point detection, network topology.
\end{keywords}

\section{Introduction}\label{sec:intro}

Networks or graphs have emerged as effective tools to understand and summarize complex systems across multiple domains of knowledge~\cite{strogatz2001, newman2010, jackson2010}.
Representing interconnected systems as graphs, where nodes correspond to agents and edges correspond to pairwise interactions between these agents, allows us to apply tools from graph theory to reveal key properties of the underlying systems such as the emergence of community structure~\cite{fortunato2010community,abbe2017community} or the existence of influential agents~\cite{pagerank1999,segarra2016stability}.

In practice, one rarely has immediate access to the whole network of interest and rather must sample it, either by directly querying the connections between pairs of nodes in a strategic fashion~\cite{ahmed2013sampling}, or performing network topology inference~\cite{mateos2019connecting,segarra2017network,shafipour2017network, dong2016learning,kalofolias2016learn}.
Once a network has been constructed, it can be used to gain insights about the system being modeled through a variety of downstream tasks.
However, if the network is subject to change, \eg{}, due to some external force influencing the propensity for connections between particular nodes, then it is in the user's interest to resample the network.
Since constructing a network can be a burdensome task~\cite{mateos2019connecting}, it is important to minimize the resampling frequency.

Although it is often infeasible to constantly resample a network to assess whether its topology has changed, we frequently have access to data supported on the nodes of the graph.
For instance, we might not be able to frequently query the connections within a social group but we can observe the preferences and actions of individuals.
Moreover, this data is typically regularized by the underlying graph structure, \eg{}, the data might be smooth on the underlying graph~\cite{kalofolias2016learn}.
In our social group example, the preferences of individuals are partially driven by the unobserved social connections.
Thus, if a nominal graph is known, a sequence of sampled data on the nodes should reflect this nominal topology up to the point where the structure changes.
At that point, the relationship between the nodal data and the nominal graph structure will deteriorate.

Working in the framework of graph signal processing, we aim to detect changes in the graph structure underlying a sequence of graph signals.
By modeling nodal data as graph signals output by graph filters, we can extract dominant features of the underlying graph solely by observing the signals.
We then leverage domain knowledge on the graph change model to propose a CUSUM-based algorithm for change-point detection.

\parheading{Related work.}
The problem of detecting changes in a sequence of graphs has been studied from several perspectives.
When a sequence of random graphs is directly available for computation, spectral methods have been designed for detecting changes in the underlying random graph model~\cite{zhang2020online,he2018sequential}.
However, these methods do not consider the case where only data on the nodes of a graph is available.
A different body of work has focused on the detection of changes in the distribution of signals on a graph.
For instance, \cite{sharpnack2015detecting,ferrari2019distributed} study the detection of a mean change in a sequence of graph signals and \cite{ferrari2020non} focuses on the change of the distribution of graph signals by leveraging the community structure of the underlying graph.
However, these approaches do not aim to detect changes in graph topology but, rather, they detect changes in the model of the data supported on a static graph.

Most related to our work, detecting changes in graph topology from graph signals has also been studied.
In~\cite{isufi2018blind}, matched subspace detectors are considered for the non-sequential case, where assumptions of bandlimitedness in the graph Fourier domain are leveraged.
Similarly, in~\cite{chepuri2016subgraph}, the generating process for the data is assumed to be known, unlike our work here which only leverages the principal eigenvectors of the data, instead of a precise distribution.

\parheading{Contributions.}
Our contributions in this paper are twofold: \\
i) We present a CUSUM algorithm for network topology change-point detection that only depends on the observation of graph signals and that can incorporate prior knowledge about the nature of the change.\\
ii) Through numerical experiments, we demonstrate the utility of even crude knowledge of the post-change graph for denoising observed data and improving detection performance.

\section{Notation and background}\label{sec:bg}

\parheading{Notation.}
The notation $[n]$ refers to the set of positive integers $\{1,2,\ldots,n\}$.
We refer to matrices using bold uppercase letters, \eg{} $\mathbf{A,B,C}$, and to (column) vectors with bold lowercase letters, \eg{} $\mathbf{v,w,x}$.
Entries of matrix $\A$ are indicated by $A_{ij}$ while those of vector $\x$ are denoted by $x_i$.
For clarity, we alternatively use the notation $[\x]_i=x_i$.
The $\ell_2$-norm of a vector is denoted by $\|\cdot\|_2$.

\parheading{Graphs and graph matrices.}
A \emph{graph} is a finite set of nodes, coupled with a set of edges connecting those nodes.
That is, for a set of $n$ nodes denoted by $\Nodes$, the set of edges $\Edges\subseteq\Nodes\times\Nodes$ forms the graph $\Graph=(\Nodes,\Edges)$.
Typically, we endow the set of nodes with arbitrarily ordered integer labels, saying that $\Nodes=[n]$.
This representation allows us to represent the graph with the \emph{adjacency matrix} $\A\in\R^{n\times n}$, where $A_{ij}=1$ if $(i,j)\in\Edges$, taking value $0$ otherwise.

In this paper, we focus on \emph{undirected} graphs where $(i,j)\in\Edges$ implies that $(j,i)\in\Edges$, \ie{}, $\Edges$ is composed of unordered pairs of nodes.
Under this condition, the adjacency matrix is symmetric.
With such an adjacency matrix, we define the \emph{Laplacian matrix} as $\L=\D-\A$,
where $\D=\diag(\A{\mathbf 1})$ is the diagonal matrix of node degrees.
The adjacency matrix and the Laplacian are two examples of \emph{graph shift operators} (generically denoted by $\Shift$), which are matrices whose sparsity and symmetry patterns correspond to those of the underlying graph~\cite{EmergingFieldGSP}.

\parheading{Graph signals and graph filters.}
We model data on the nodes of a graph as a \emph{graph signal}.
A graph signal is a real-valued function supported on the nodes of a graph $x:\Nodes\to\R$ that can be conveniently represented as a vector $\x$ in $\Rn$, where $[\x]_i=x(i)$ for each $i\in[n]$.
We now define the notion of \emph{graph filters} as linear maps between graph signals.
Assuming that a given graph shift operator $\Shift$ has the eigenvalue decomposition $\Shift = \sum_i\lambda_i\v_i\v_i^\top$, a graph filter $\Filt(\Shift)$ is a real polynomial of $\Shift$, \eg{} for coefficients $\alpha_k$,
\begin{equation}\label{eq:spectral-graph-filter}
\Filt(\Shift)=\sum_{k=0}^T\alpha_k\Shift^k=\sum_{i=1}^n h\left(\lambda_i\right)\v_i\v_i^\top,
\end{equation}
where $h:\R\to\R$ is the extension of $\Filt$ to the real numbers.
Graph filters have shown to be versatile tools in modeling linear network processes~\cite{segarra2017optimal}.
Notice that graph filters preserve the eigenvectors of the underlying graph shift operator, while distorting its eigenvalues.

\section{Network change-point detection}\label{sec:cpd}

Consider a streaming sequence of graphs with a single change point.
More precisely, we have a nominal graph $\Graph_0=(\Nodes,\Edges_0)$, a post-change graph $\Graph_1=(\Nodes,\Edges_1)$, and a sequence of graphs $\{\Graph^{(t)}\}_{t=1}^\infty$ modeled as
\begin{equation}\label{eq:graph-sequence}
  \begin{array}{ll}
    \Graph^{(t)}=\Graph_0, & t<\tau, \\
    \Graph^{(t)}=\Graph_1, & t\geq\tau.
  \end{array}
\end{equation}
In~\eqref{eq:graph-sequence}, the graphs in the given sequence are equal to $\Graph_0$ before the unknown change point $t=\tau$ and, after the change point, they are equal to $\Graph_1$.

Our objective is to develop a method for detecting this change without direct observations of the graphs $\Graph^{(t)}$ or precise knowledge of the post-change graph $\Graph_1$.
Specifically, we model the observations available to us as a sequence of graph signals given by
\begin{equation}\label{eq:signal-model}
  \y^{(t)} =
  \begin{cases}
    \Filt_0(\Shift_0)\, \w^{(t)} & \text{for }\, t<\tau, \\
    \Filt_1(\Shift_1)\, \w^{(t)} & \text{for }\, t\geq\tau,
  \end{cases}
\end{equation}
where $\Shift_0$ and $\Shift_1$ are graph shift operators of the graphs $\Graph_0$ and $\Graph_1$ in~\eqref{eq:graph-sequence}, respectively. 
Furthermore, we consider the challenging case where the graph filters $\Filt_0$ and $\Filt_1$ in~\eqref{eq:signal-model} have unknown coefficients, and the unknown inputs $\w^{(t)}$ are independent random vectors drawn from a common centered distribution with identity covariance.
With this model in place, we now formally state the problem of interest.
\begin{problem}\label{prob:chgpnt}
  Given a streaming sequence of graph signals modeled by~\eqref{eq:signal-model}, detect at time $t$ whether or not $t\geq\tau$.
\end{problem}
\noindent\Cref{prob:chgpnt} prompts us to decide, in real time, whether the change in the underlying graph has already occurred or not.
This is motivated by the desire to detect graph topology changes when the correlation structure of the observed data is determined by interactions between neighboring nodes, \eg{} to detect structural changes in social networks from opinion data over time, or anomalies in sensor networks from nodal measurements.

In order to solve~\Cref{prob:chgpnt}, we must be able to extract key features of the underlying graphs from the observed signals $\y^{(t)}$.
In particular, a direct computation of the covariance $\Cy^{(t)}$ of $\y^{(t)}$ -- leveraging~\eqref{eq:spectral-graph-filter} and the identity covariance of the inputs $\w^{(t)}$ -- reveals that $\Cy^{(t)}$ preserves the eigenvectors of the underlying graph shift operators at time $t$.
More precisely, if the shift operators have eigenvalue decompositions $\Shift_0=\sum_{i}\lambda_i\v_i\v_i^\top$ and $\Shift_1=\sum_{i}\mu_i\u_i\u_i^\top$, then
\begin{equation}\label{eq:signal-covariance}
 \Cy^{(t)}=
  \begin{cases}
    \Cy^0:=\sum_{i=1}^n h_0^2(\lambda_i)\v_i\v_i^\top & \text{for }\, t<\tau, \\
    \Cy^1:=\sum_{i=1}^n h_1^2(\mu_i)\u_i\u_i^\top & \text{for }\, t\geq\tau,
  \end{cases}
\end{equation}
where $h_0$ and $h_1$ are the respective extensions of the matrix polynomials $\Filt_0$ and $\Filt_1$ to the real numbers.
It can be seen from~\eqref{eq:signal-covariance} that we can estimate the eigenvectors of the shift operator at time $t$ by taking the sample covariance matrix of graph signals near that time.

Following this direction and in order to formulate a tractable hypothesis test for the solution of \Cref{prob:chgpnt}, we restrict our view to the \emph{dominant subspaces} of the covariance matrices $\Cy^0$ and $\Cy^1$ in~\eqref{eq:signal-covariance}.
Thus, we translate our assumed knowledge of $\Graph_0$ and lack of precise knowledge of $\Graph_1$ into statements about these dominant subspaces.
In particular, we assume that the leading subspace of $\Cy^0$ is known and denoted by $\U_0$, and that the leading subspace of $\Cy^1$ is known to be an element of \emph{a family of subspaces} $\{\U_1(\gamma)\}_{\gamma\in\Gamma}$, for some set $\Gamma$.
Notice that this assumption of a parameterized post-change subspace is realistic, \eg{}, in a social network under a polarization effect, $\gamma$ could dictate how the nodes split into communities, where $\U_1(\gamma)$ is the dominant subspace of a stochastic block model parameterized by $\gamma$~\cite{abbe2017community}.
For simplicity, we assume that all the dominant subspaces considered are $k$-dimensional.
Then, by partitioning the observed sequence of graph signals $\{\y^{(t)}\}_{t=1}^\infty$ into consecutive blocks of size $b$, we form estimates of the current dominant subspace by taking the top $k$ eigenvectors of the sample covariance matrix of each block. 
We denote these subspaces by $\Vhat^{(\ell)}$, where $\ell$ indicates the block under consideration.
This naturally leads to the following hypothesis test for change-point detection
\begin{equation}\label{eq:hypothesis-test}
  \begin{array}{ll}
    H_0: & \V^{(t)} = \U_0, \\
    H_1: & \V^{(t)} = \U_1(\gamma)\text{ for some }\gamma\in\Gamma,
  \end{array}
\end{equation}
where $\V^{(t)}$ is the true dominant subspace of the covariance matrix at time $t$, estimated by the corresponding $\Vhat^{(\ell)}$.
The form of~\eqref{eq:hypothesis-test} is tractable in the setting where the graphs are only observed via data on their nodes, allowing us to leverage sequential subspace detection methods~\cite{xie2018first,jiao2018subspace}.

\begin{algorithm}[tb]
	\caption{Network change-point detection with prior parameterized post-change subspaces}\label{alg:chgpnt}
	\begin{algorithmic}[1]
		\State \textbf{Input:} Sequence of estimated dominant subspaces $\{\Vhat^{(\ell)}\}_{\ell=1}^\infty$
		\State \textbf{Parameters:} Nominal subspace $\U_0$, family of post-change subspaces $\{\U_1(\gamma)\}_{\gamma\in\Gamma}$
		\State \textbf{Parameters:} CUSUM correction parameter $c$, decision threshold $\eta$
		\State Let $d(\V,\U)=\|\sin\Theta(\V,\U)\|_F$ for subspaces $\V,\U$
		\State $S^{(0)}\gets 0$, $\ell\gets 1$
		\While{$S^{(\ell-1)}<\eta$}
		\State $\widehat{\gamma}^{(\ell)}\gets\argmin_{\gamma\in\Gamma} d(\U_1(\gamma),\Vhat^{(\ell)})$
		\State $L^{(\ell)}\gets d(\U_0,\U_1(\widehat{\gamma}^{(\ell)}))-d(\U_1(\widehat{\gamma}^{(\ell)},\Vhat^{(\ell)}))-c$
		\State $S^{(\ell)}\gets\max(0,S^{(\ell-1)}+L^{(\ell)})$
                \State $\ell\gets\ell+1$
		\EndWhile
		\State \textbf{Output:} change detected
	\end{algorithmic}
\end{algorithm}

With this formulation in place, we propose~\cref{alg:chgpnt} to test the hypothesis~\eqref{eq:hypothesis-test} by applying a CUSUM-based~\cite{page1954continuous,pignatiello1990comparisons} method to our sequence of estimated subspaces $\{\Vhat^{(\ell)}\}_{\ell=1}^\infty$.
For each  $\ell$ we determine the subspace in $\{\U_1(\gamma)\}_{\gamma\in\Gamma}$ that is closest to $\Vhat^{(\ell)}$, where we measure the distance between subspaces using the Frobenius sin $\Theta$ distance~\cite{wong1967differential} (see line 4 in~\cref{alg:chgpnt}).
We denote this closest subspace as $\U_1(\widehat{\gamma}^{(\ell)})$. 
We then compare the distances from the observed $\Vhat^{(\ell)}$ to $\U_1(\widehat{\gamma}^{(\ell)})$ and from the nominal $\U_0$ to $\U_1(\widehat{\gamma}^{(\ell)})$ within a conventional CUSUM framework.
In a nutshell, if the distance between the nominal subspace and $\U_1(\widehat{\gamma}^{(\ell)})$ is significantly larger -- where the level of significance is determined by the correction parameter $c$ -- than the distance between the observed subspace and $\U_1(\widehat{\gamma}^{(\ell)})$, we have reason to suspect that a topology change has occurred and we increase the value of our CUSUM statistic.
This approach is inspired by the approach of~\cite{chen2018new}, where the shift in the mean of a multivariate Gaussian random variable is assumed to be in a known direction, and the most likely scale of that shift is estimated at each step.

\begin{remark}\label{rem:flex}
The choice of post-change subspaces $\{\U_1(\gamma)\}_{\gamma\in\Gamma}$ confers a high level of flexibility to~\Cref{alg:chgpnt}.
Indeed, in the absence of any information about the post-change graph $\Graph_1$, one might adopt the most flexible setting, where $\{\U_1(\gamma)\}_{\gamma\in\Gamma}$ contains \emph{all} $k$-dimensional subspaces of $\Rn$.
Under this setting, $\Vhat^{(\ell)}=\U_1(\widehat{\gamma}^{(\ell)})$ for all $\ell$, so that~\cref{alg:chgpnt} only relies on the distance between $\U_0$ and $\Vhat^{(\ell)}$ at each step.
That is, $L^{(\ell)}=d(\U_0,\Vhat^{(\ell)})-c$, thus boiling down to a more conventional CUSUM statistic for detecting deviations in any direction from an expected behavior.
\end{remark}

\parheading{The effect of the correction parameter $c$.}
The performance of~\cref{alg:chgpnt} is dictated by $c$ and the quality of the estimated dominant subspaces $\{\Vhat^{(\ell)}\}_{\ell=1}^\infty$.
Given the difficulty of precisely characterizing the distribution of the estimated subspaces, \cref{alg:chgpnt} deviates from the standard formulation of CUSUM for Gaussian random variables based on likelihood (or generalized likelihood) ratios~\cite{healy1987note}.
Nonetheless, we follow the provably-valid strategy of substituting the likelihood ratio with a general function with negative expected value in the nominal state and positive expected value after the change~\cite{oskiper2005quickest}.
For this to hold, we require the following condition relating our estimates, our candidate post-change subspaces, and the correction parameter $c$,
\begin{equation}\label{eq:corr-bounds}
  \begin{split}
    \E_0\left[d_{\U_0,\U_1(\widehat{\gamma}^{(\ell)}), \Vhat^{(\ell)}}\right] < c 
    < \E_1\left[d_{\U_0,\U_1(\widehat{\gamma}^{(\ell)}), \Vhat^{(\ell)}}\right],
\end{split}
\end{equation}
where $d_{\U_0,\U_1(\widehat{\gamma}^{(\ell)}), \Vhat^{(\ell)}}:= d(\U_0,\U_1(\widehat{\gamma}^{(\ell)}))-d(\Vhat^{(\ell)},\U_1(\widehat{\gamma}^{(\ell)}))$, and $\E_0$ and $\E_1$ refer to the nominal and post-change expectations, respectively.
\begin{figure}[t]
	\centering
	\def\svgwidth{0.5\linewidth}
	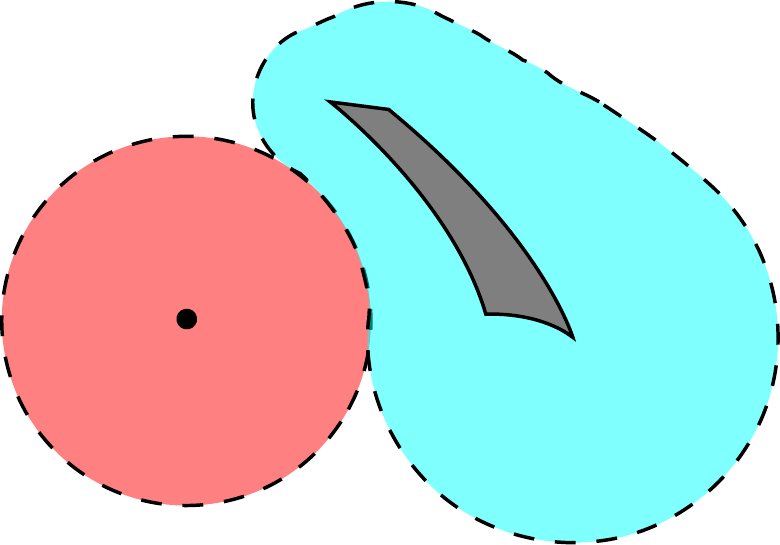
	\caption{
          Geometry of $L^{(\ell)}$ for two different choices of the post-change subspaces $\{\U_1(\gamma)\}_{\gamma\in\Gamma}$.
		\vspace{-0.5cm}
	}
	\label{fig:regions}
\end{figure}
We illustrate the geometry of this requirement in~\cref{fig:regions} for the $k=1$ dimensional case.
The red circle around $\U_0$ indicates the set of $1$-dimensional estimated subspaces such that $d(\U_0,\Vhat^{(\ell)})-c<0$, as in the case where $\{\U_1(\gamma)\}_{\gamma\in\Gamma}$ contains all $1$-dimensional subspaces of $\Rn$ (cf. Remark~\ref{rem:flex}).
In this setting, all estimated subspaces beyond this radius obey $d(\U_0,\Vhat^{(\ell)})-c>0$.
However, when there is prior information about the post-change subspaces (as in the gray region $\{\U_1(\gamma)\}$), only the cyan region yields $d(\U_0,\U_1(\widehat{\gamma}^{(\ell)})) - d( \Vhat^{(\ell)}, \U_1(\widehat{\gamma}^{(\ell)}) )-c>0$ in order to contribute to the CUSUM statistic.
In a sense, only subspace estimates that are pointed in the right direction from $\U_0$ to $\{\U_1(\gamma)\}_{\gamma\in\Gamma}$ contribute to the CUSUM statistic, rather than all subspaces beyond the open ball drawn in red.
Intuitively, fulfilling the requirement in~\eqref{eq:corr-bounds} amounts to choosing the parameter $c$ such that most of the estimated nominal subspaces lie in the open ball around $\U_0$ and most of the estimated post-change subspaces lie in the region around $\{\U_1(\gamma)\}_{\gamma\in\Gamma}$.

\section{Experiments}\label{sec:exp}

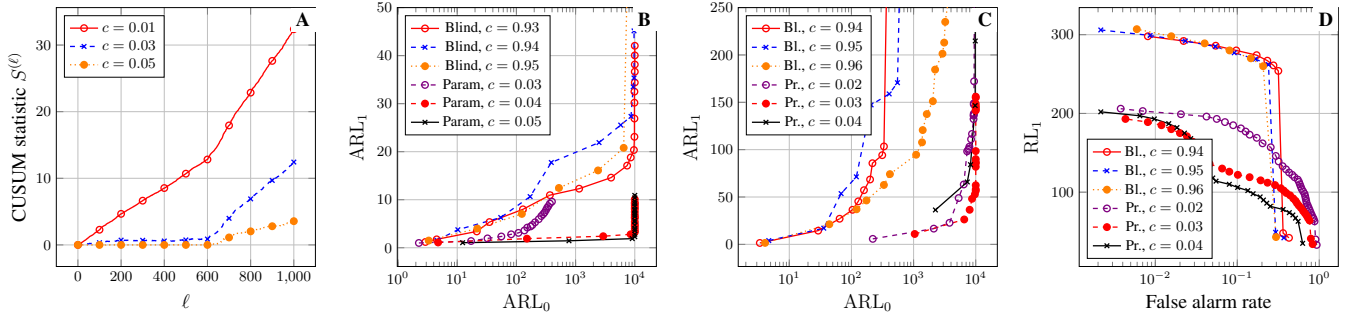
\begin{figure*}[t!]
  \centering
  \resizebox{\linewidth}{!}{\input{fig/plots.tikz}}
  \caption{
    \normalsize
    \textbf{(A)} CUSUM statistic for detecting the change of an ER graph into a BA graph.
    \textbf{(B)} Average run length before ($\mathrm{ARL}_0$) and after ($\mathrm{ARL}_1$) the change for detection of a BA graph from an ER graph.
    \textbf{(C)} Average run length before and after the change for detection of an emerging community.
    \textbf{(D)} Detection of change points in the market from mentions on Twitter. False alarm rate for detection of changes before March 9th 2015 compared to the run length after the morning of March 9th.
    \vspace{-0.5cm}
  }
  \label{fig:exp}
\end{figure*}

\cref{alg:chgpnt} relies on a parameterization of the dominant subspace of the post-change graph in terms of $\gamma\in\Gamma$.
In practice, however, the spectral properties of the post-change graph may only be modeled roughly.
In this section, we demonstrate how even the use of an \emph{approximate} parameterization of the post-change subspace can improve CUSUM-based subspace change-point detection algorithms.
In the synthetic experiments, both graph filters in~\eqref{eq:signal-model} are equal to the adjacency matrix squared, \ie{} $\Filt_0(\A_0)=\A_0^2, \Filt_1(\A_1)=\A_1^2$, with $\w$ being drawn from a standard normal multivariate Gaussian distribution.
Under this model, the dominant subspaces of the nominal and post-change covariance matrices respectively correspond to the eigenvectors of $\A_0$ and $\A_1$ with the highest-magnitude eigenvalues.

\parheading{Detecting concentrations of centrality.}
We consider a model where the nominal graph is an \ER{} (ER) graph on $n=100$ nodes with density parameter $p=2\log n/n$.
The post-change graph is a randomly drawn \BA{} (BA) network on the same set of nodes with parameters $m_0=m=1$.
In terms of eigenvector centralities, one expects the pre-change centrality to be flat across nodes, in accordance with the expectation of the ER graph, and the post-change centrality to be concentrated in a few nodes, in accordance with the power-law structure of the BA random graph model~\cite{farkas2001spectra}.

We focus on single dimensional leading subspaces for our change-point detection and we do not assume that the high-centrality nodes in the post-change graph are known.
Rather, we crudely approximate the post-change structure of the leading eigenvector (centrality) with a spike: a discrete delta function $\u_1(\gamma)=\delta_\gamma$, for $\gamma\in[n]$.
Evaluating the distances in~\cref{alg:chgpnt} boils down to finding the maximum magnitude element of the estimated leading eigenvector, and then comparing to the same element in the nominal eigenvector.
Formally, if we have that $\widehat{\gamma}^{(\ell)} = \argmax_{\gamma\in[n]} |[\vhat^{(\ell)}]_\gamma|$ then
\begin{equation}\label{eq:exp:spike:distances}
  \begin{split}
    d\big(\vhat^{(\ell)},\u_1(\widehat{\gamma}^{(\ell)})\big)&=\sqrt{1-\|\vhat^{(\ell)}\|_\infty^2},\\
    d\big(\u_0,\u_1(\widehat{\gamma}^{(\ell)})\big)&=\sqrt{1-[\u_0]_{\widehat{\gamma}^{(\ell)}}^2},
  \end{split}
\end{equation}
where $\vhat^{(\ell)}$ is the estimated leading eigenvector, $\u_1(\gamma)$ is a vector taking value $1$ at index $\gamma$ and $0$ elsewhere, and $\u_0$ is the leading eigenvector of the nominal ER graph.

We illustrate the case where the block size is $b=1$, \ie{}, the estimate of the leading eigenvector $\vhat^{(\ell)}$ is given simply by the observed graph signal at time $t = \ell$, scaled to have unit norm.
The CUSUM statistics $S^{(\ell)}$ of this experiment are shown in~\letterref{fig:exp}{A}, for varying correction parameters $c$ on a sequence of $1000$ signals with a change point at $\tau=600$.
When the correction parameter is small ($c=0.01$), the left inequality in~\eqref{eq:corr-bounds} is not satisfied.
This leads to an increase in the CUSUM statistic even before the change point.
However, when $c$ is sufficiently large, the nature of the CUSUM statistic is apparent.
It remains close to zero in the nominal state, and then increases with slope inversely proportional to $c$ after the change point.

We compare our method using the parameterization of the post-change subspace with a delta function with a method using no knowledge of the post-change subspace, where $\{\u_1(\gamma)\}_{\gamma\in\Gamma}$ is the set of all 1-dimensional subspaces of $\Rn$, as discussed in~\cref{rem:flex}.
We generate 10000 signals in the nominal state, as well as 10000 post-change signals, and compare both methods for varying correction parameters $c$, plotting the expected nominal and post-change run lengths for a range of decision thresholds $\eta$, as in~\cite[Section 5]{jiao2018subspace}.
As illustrated in~\letterref{fig:exp}{B}, for a fixed average run length in the nominal state ($\mathrm{ARL}_0$), the parameterized approach detects the change more quickly ($\mathrm{ARL}_1$) than the blind approach, where no prior information is used.
Although the post-change subspace will never take the form of a delta function, this result shows that even a rough estimate of how the post-change graph is structured can be leveraged to improve change-point detection algorithms.

\parheading{Detecting emerging communities.}
To illustrate our approach on a more complex subspace problem, let the nominal graph again be an ER graph of $n=100$ nodes with density parameter $p=2\log n/n$.
For some unknown cutoff $1\leq n_0\leq n$, we model the post-change graph as identical to the nominal graph but having the edges $(i,j)$ for $i,j\leq n_0$ redrawn with density parameter $q=5p$.

This models the emergence of a community dictated by strong adherence to a particular ideology, \eg{}, a faction of the most politically extreme individuals in a social network forming a dense community, where the sorted labeling of the nodes implicitly places nodes on the political spectrum.
In this setting, our nominal subspace hypothesis compares the estimated subspace to the leading $k=2$ eigenvectors of the nominal graph, while the post-change hypothesis compares the estimated subspace to the $k=2$ leading eigenvectors of the expected post-change adjacency matrix, parameterized by $\gamma=n_0$.
We consider signal blocks of size $b=50$ in estimating the leading subspaces.
As shown in~\letterref{fig:exp}{C}, using the parameterized (Pr.) model of the post-change subspace is effective in denoising the observed signals, yielding $\mathrm{ARL}_0/\mathrm{ARL}_1$ curves that are below those of the case where no knowledge of the post-change subspace is assumed (Bl.).

\parheading{Detecting events in Twitter data.}
To illustrate the utility of this approach on real data, we consider a dataset recording daily mentions of 10 companies on Twitter, where the signals count the number of mentions of each company every 5 minutes.
We determine the nominal dominant eigenvector from the sample covariance of the first 4 days of data, where we have implicitly assumed the existence of an underlying network structure between these companies.
We aim to detect sudden events related to public activity of a company over the next 10 days.
Our post-change hypothesis is similar to our first experiment, where we model the post-change dominant subspace as a discrete delta function.
To estimate the subspaces $\vhat^{(\ell)}$, we use a sliding window of width $b=36$ samples, corresponding to 3-hour periods of time.
As a ground-truth change point, we consider the morning of March 9th, 2015, which corresponds to a special event from Apple.
We compare our parameterization with a delta function to the case where the post-change subspace is not parameterized.
In~\letterref{fig:exp}{D}, we compare the false alarm rate, defined as the proportion of time before the ground-truth change point that the CUSUM statistic $S^{(\ell)}$ is above the threshold, to the number of samples after the change point needed for detection.
By leveraging our model of the post-change subspace, we achieve earlier detection ($\mathrm{RL}_1$) than the case where no model is used.
This shows that~\cref{alg:chgpnt} has applications beyond the case where a precise network is known, since it only relies on having a model for the shape of the post-change dominant eigenvectors in the data.

\section{Conclusion}\label{sec:conc}

We have considered a graph change-point detection problem, where we detect a change in the graph without directly observing its topology.
Instead, we have access to data on the nodes, which is shaped by the underlying graph via some unknown network process.
Through the lens of graph signal processing, we frame this as a subspace detection problem, and then propose an algorithm allowing one to incorporate prior knowledge on the structure of the post-change graph.
We demonstrate how this prior domain knowledge can improve CUSUM-based detection algorithms on synthetic and real data, even when the precise post-change structure is not known.


\vfill\pagebreak

\bibliographystyle{IEEEbib}
\bibliography{refs}

\end{document}

%% file: fig/regions.pdf_tex
\begingroup%
  \makeatletter%
  \providecommand\color[2][]{%
    \errmessage{(Inkscape) Color is used for the text in Inkscape, but the package 'color.sty' is not loaded}%
    \renewcommand\color[2][]{}%
  }%
  \providecommand\transparent[1]{%
    \errmessage{(Inkscape) Transparency is used (non-zero) for the text in Inkscape, but the package 'transparent.sty' is not loaded}%
    \renewcommand\transparent[1]{}%
  }%
  \providecommand\rotatebox[2]{#2}%
  \newcommand*\fsize{\dimexpr\f@size pt\relax}%
  \newcommand*\lineheight[1]{\fontsize{\fsize}{#1\fsize}\selectfont}%
  \ifx\svgwidth\undefined%
    \setlength{\unitlength}{224.59099919bp}%
    \ifx\svgscale\undefined%
      \relax%
    \else%
      \setlength{\unitlength}{\unitlength * \real{\svgscale}}%
    \fi%
  \else%
    \setlength{\unitlength}{\svgwidth}%
  \fi%
  \global\let\svgwidth\undefined%
  \global\let\svgscale\undefined%
  \makeatother%
  \begin{picture}(1,0.69808941)%
    \lineheight{1}%
    \setlength\tabcolsep{0pt}%
    \put(0,0){\includegraphics[width=\unitlength,page=1]{fig/regions.pdf}}%
    \put(0.15,0.24){\makebox(0,0)[lt]{\lineheight{1.25}\smash{\begin{tabular}[t]{l}{$\U_0$}\end{tabular}}}}%
    \put(0.60,0.19){\makebox(0,0)[lt]{\lineheight{1.25}\smash{\begin{tabular}[t]{l}{$\{\U_1(\gamma)\}$}\end{tabular}}}}%
  \end{picture}%
\endgroup%

%% file: fig/plots.tikz
\begin{tikzpicture}

  \begin{groupplot}[
    group style={
      group name=myplots,
      group size=4 by 1,
      horizontal sep=1.75cm,
    },
    every axis label/.append style={
      font=\large,
    },
    tick align=inside,
    tick pos=both,
    xmajorgrids,
    xticklabel style={
      /pgf/number format/fixed,
      /pgf/number format/precision=2,
    },
    ymajorgrids,
    yticklabel style={
      /pgf/number format/fixed,
      /pgf/number format/precision=2,
    },
    scaled y ticks=false,
    width=0.4\textwidth,
    height=0.4\textwidth,
    ]

    \nextgroupplot[
    legend pos=north west,
    xlabel={$\ell$},
    ylabel={CUSUM statistic $S^{(\ell)}$},
    mark repeat=100,
    ]
    \addplot table[x=Index, y=0.01, col sep=comma] {./output/spike-cusum/nc_ss_cusum.csv};
    \addlegendentry{$c=0.01$}
    \addplot table[x=Index, y=0.03, col sep=comma] {./output/spike-cusum/nc_ss_cusum.csv};
    \addlegendentry{$c=0.03$}
    \addplot table[x=Index, y=0.05, col sep=comma] {./output/spike-cusum/nc_ss_cusum.csv};
    \addlegendentry{$c=0.05$}

    \nextgroupplot[
    legend pos=north west,
    xmode=log,
    ymode=normal,
    ymax=50,
    xlabel={$\mathrm{ARL}_0$},
    ylabel={$\mathrm{ARL}_1$},
    ]
    \addplot table[x=ARL, y=EDD, col sep=comma] {./output/spike-eddarl/nc_arl_edd_blind_0.93.csv};
    \addlegendentry{Blind, $c=0.93$}
    \addplot table[x=ARL, y=EDD, col sep=comma] {./output/spike-eddarl/nc_arl_edd_blind_0.94.csv};
    \addlegendentry{Blind, $c=0.94$}
    \addplot table[x=ARL, y=EDD, col sep=comma] {./output/spike-eddarl/nc_arl_edd_blind_0.95.csv};
    \addlegendentry{Blind, $c=0.95$}
    \addplot table[x=ARL, y=EDD, col sep=comma] {./output/spike-eddarl/nc_arl_edd_param_0.03.csv};
    \addlegendentry{Param, $c=0.03$}
    \addplot table[x=ARL, y=EDD, col sep=comma] {./output/spike-eddarl/nc_arl_edd_param_0.04.csv};
    \addlegendentry{Param, $c=0.04$}
    \addplot table[x=ARL, y=EDD, col sep=comma] {./output/spike-eddarl/nc_arl_edd_param_0.05.csv};
    \addlegendentry{Param, $c=0.05$}

    \nextgroupplot[
    legend pos=north west,
    xmode=log,
    ymode=normal,
    ymax=250,
    xlabel={$\mathrm{ARL}_0$},
    ylabel={$\mathrm{ARL}_1$},
    ]
    \addplot table[x=ARL, y=EDD, col sep=comma] {./output/comm-eddarl/ef_arl_edd_blind_0.94.csv};
    \addlegendentry{Bl., $c=0.94$}
    \addplot table[x=ARL, y=EDD, col sep=comma] {./output/comm-eddarl/ef_arl_edd_blind_0.95.csv};
    \addlegendentry{Bl., $c=0.95$}
    \addplot table[x=ARL, y=EDD, col sep=comma] {./output/comm-eddarl/ef_arl_edd_blind_0.96.csv};
    \addlegendentry{Bl., $c=0.96$}
    \addplot table[x=ARL, y=EDD, col sep=comma] {./output/comm-eddarl/ef_arl_edd_param_0.02.csv};
    \addlegendentry{Pr., $c=0.02$}
    \addplot table[x=ARL, y=EDD, col sep=comma] {./output/comm-eddarl/ef_arl_edd_param_0.03.csv};
    \addlegendentry{Pr., $c=0.03$}
    \addplot table[x=ARL, y=EDD, col sep=comma] {./output/comm-eddarl/ef_arl_edd_param_0.04.csv};
    \addlegendentry{Pr., $c=0.04$}

    \nextgroupplot[
    legend pos=south west,
    xmode=log,
    ymode=normal,
    xlabel={False alarm rate},
    ylabel={$\mathrm{RL}_1$},
    ]
    \addplot table[x=FA Rate, y=Detection Delay, col sep=comma] {./output/twitter-eddarl/tw_fa_dd_blind_0.86.csv};
    \addlegendentry{Bl., $c=0.94$}
    \addplot table[x=FA Rate, y=Detection Delay, col sep=comma] {./output/twitter-eddarl/tw_fa_dd_blind_0.88.csv};
    \addlegendentry{Bl., $c=0.95$}
    \addplot table[x=FA Rate, y=Detection Delay, col sep=comma] {./output/twitter-eddarl/tw_fa_dd_blind_0.9.csv};
    \addlegendentry{Bl., $c=0.96$}
    \addplot table[x=FA Rate, y=Detection Delay, col sep=comma] {./output/twitter-eddarl/tw_fa_dd_param_0.56.csv};
    \addlegendentry{Pr., $c=0.02$}
    \addplot table[x=FA Rate, y=Detection Delay, col sep=comma] {./output/twitter-eddarl/tw_fa_dd_param_0.58.csv};
    \addlegendentry{Pr., $c=0.03$}
    \addplot table[x=FA Rate, y=Detection Delay, col sep=comma] {./output/twitter-eddarl/tw_fa_dd_param_0.62.csv};
    \addlegendentry{Pr., $c=0.04$}

  \end{groupplot}
  
  \node[below left,fill=white] at (myplots c1r1.north east) {\large\textbf{A}};
  \node[below left,fill=white] at (myplots c2r1.north east) {\large\textbf{B}};
  \node[below left,fill=white] at (myplots c3r1.north east) {\large\textbf{C}};
  \node[below left,fill=white] at (myplots c4r1.north east) {\large\textbf{D}};
  
\end{tikzpicture}
